\documentclass[conference]{IEEEtran}
\IEEEoverridecommandlockouts
\usepackage{cite}
\usepackage{amsmath,amssymb,amsfonts}
\usepackage{algorithmic}
\usepackage{graphicx}
\usepackage{textcomp}
\usepackage{xcolor}
\usepackage{multirow}
\usepackage{balance}
\def\BibTeX{{\rm B\kern-.05em{\sc i\kern-.025em b}\kern-.08em
    T\kern-.1667em\lower.7ex\hbox{E}\kern-.125emX}}
\begin{document}

\title{Kitchen Food Waste Image Segmentation and Classification for Compost Nutrients Estimation\\

\thanks{The work is partly supported by The Natural Sciences and Engineering Research Council of Canada (NSERC) and Mitacs Accelerate Program.\\$^\dagger$ These authors contributed equally to this work.}
}

\author{Raiyan Rahman\textsuperscript{\rm 1}$^\dagger$, 	Mohsena Chowdhury\textsuperscript{\rm 1}$^\dagger$, 
Yueyang Tang\textsuperscript{\rm 2}, Huayi Gao\textsuperscript{\rm 2}, George Yin\textsuperscript{\rm 2}, Guanghui Wang\textsuperscript{\rm 1}$^*$\\
\textsuperscript{\rm 1}Department of Computer Science, Toronto Metropolitan University, 350 Victoria Street
Toronto, ON M5B 2K3 \\ 
\textsuperscript{\rm 2}VCycene Inc. 3600 Steeles Ave. E. Markham, ON L3R 9Z7 \\ 
\tt\small \{raiyan.rahman, mohsena.chowdhury\}@torontomu.ca, \{cris, huayi,george\}@virgohome.io \\ \tt\small {wangcs@torontomu.ca (* corresponding author)}
}

\maketitle

\begin{abstract}
The escalating global concern over extensive food wastage necessitates innovative solutions to foster a net-zero lifestyle and reduce emissions. The LILA home composter presents a convenient means of recycling kitchen scraps and daily food waste into nutrient-rich, high-quality compost. To capture the nutritional information of the produced compost, we have created and annotated a large high-resolution image dataset of kitchen food waste with segmentation masks of 19 nutrition-rich categories. Leveraging this dataset, we benchmarked four state-of-the-art semantic segmentation models on food waste segmentation, contributing to the assessment of compost quality of Nitrogen, Phosphorus, or Potassium. The experiments demonstrate promising results of using segmentation models to discern food waste produced in our daily lives. Based on the experiments, SegFormer, utilizing MIT-B5 backbone, yields the best performance with a mean Intersection over Union (mIoU) of 67.09. Class-based results are also provided to facilitate further analysis of different food waste classes.
\end{abstract}


\section{Introduction}
Food waste has significant implications for our lives, the environment, economic consequences, and the global community. 
Taking Canada as an example, it is reported that about 396 kilograms of food annually are wasted or lost per capita, making it one of the top food waste generators in the world \cite{b1}. In Canada, more than 32\% of our methane (CH4) production is contributed by food waste in landfills \cite{b2}. Turning food waste directly into organic compost is a great way to incentivize this transition into a net-zero sustainable lifestyle for each Canadian household while contributing to cutting down our carbon emissions and meeting the goals our government has set for the Paris Agreement, supporting national efforts to meet environmental targets and contribute to a more sustainable future.

\begin{figure}
  \centering
  \includegraphics[width=0.95\columnwidth]{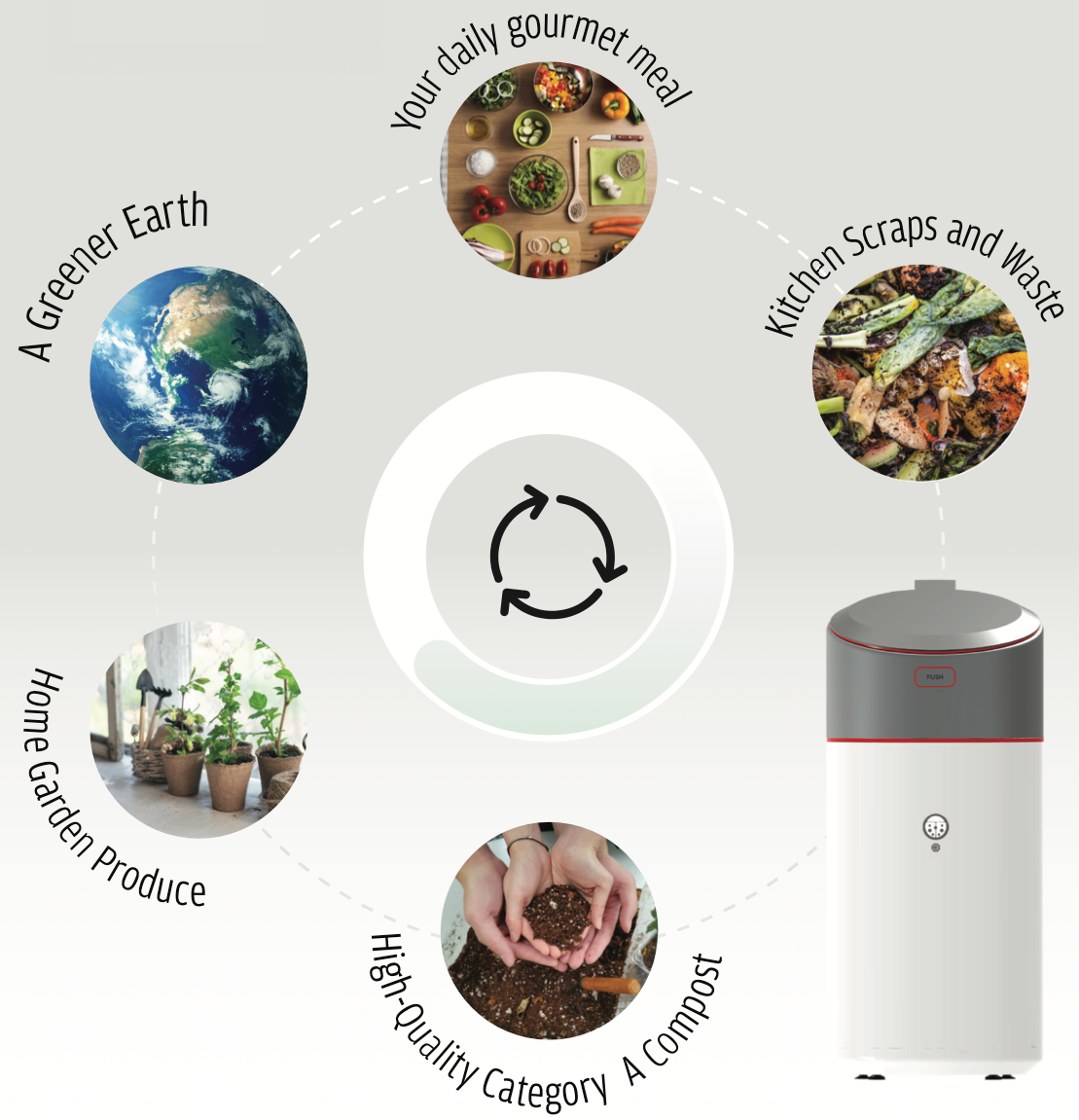}
  \caption{The home composter would conveniently allow households to recycle their kitchen scraps and daily food waste into high-quality compost rich in nutrients. This paper aims to study the feasibility of using semantic segmentation techniques to identify food waste classes, enabling the capture of nutritional information from food waste.}
  \label{fig:lilacomposter}
\end{figure}

To tackle this intricate issue, we have engineered a home composter named LILA, designed to seamlessly, efficiently, and inconspicuously transform kitchen-generated food waste into organic compost, as depicted in Fig. \ref{fig:lilacomposter}. LILA excels in consistently sorting waste, conditioning, and converting food waste into fully mature organic compost. This system offers a more effective and odor-free solution for recycling food waste, and the transformation of food waste to organic compost presents a promising solution, helping to mitigate waste generation and greenhouse gas emissions, and fostering a more sustainable and environmentally conscious lifestyle.

The primary objective of this paper is to empower households to capture crucial nutritional information, specifically the NPK (Nitrogen, Phosphorus, and Potassium) values, associated with the compost generated by the composter. This information equips users with the knowledge to judiciously utilize the compost, fostering enhanced agricultural yields and the remediation of contaminated soil. Since various types of food waste, after the composting process, yield distinct NPK values, a key aspect is the automatic recognition of the type of food waste and its corresponding quantity. 
To achieve this, we propose to utilize computer vision techniques for the segmentation and recognition of different classes of food waste based on their images. This study endeavors to explore the effectiveness and feasibility of employing semantic segmentation techniques for this specific task. 

The main contributions of this paper are summarized below.

\begin{table*}
    \centering
    \caption{Table detailing the food waste classes present in the dataset and their corresponding representation as well as nutritional information.}
    \begin{tabular}{|| c | c | c | c | c | c ||} 
         \hline
         \textbf{Class} & \textbf{\# Images} & \textbf{Nutrition Information} & \textbf{Nitrogen, N (mg)} & \textbf{Phosphorus, P (mg)} & \textbf{Potassium, K (mg)} \\ [0.5ex]
         \hline
         Banana Skin & 485 & Balanced & 443.75 & 100 & 420\\
         \hline
         Egg Shell & 581 & Balanced & 350 & 160 & 150\\
         \hline
         Lettuce Leaf & 410 & Nutrition-light & 180 & 27 & 91\\
         \hline
         Hard Bread & 261 & Nutrition-rich & 1970 & 212 & 250\\
         \hline
         Cooked Meat & 201 & Nutrition-rich & 3260 & 280 & 476\\
         \hline
         Onion Skin & 493 & Balanced & 431.25 & 300.36 & 161.20\\
         \hline
         Potato Skin & 232 & Nutrition-rich & 3152 & 262.20 & 287.14\\
         \hline
         Apple Core & 212 & Nutrition-light & 4.2 & 72 & 95\\
         \hline
         Orange & 107 & Nutrition-light & 140 & 23 & 166\\
         \hline
         Waffle & 43 & Nutrition-rich, Nitrogen-rich & 1510 & 254 & 217\\
         \hline
         Apple Peel & 164 & Nutrition-median, Potassium-rich & 12.5 & 12 & 257.57\\
         \hline
         Corn Leaves & 44 & Nutrition-light & 28 & 1.5 & 16.6\\
         \hline
         Cucumber & 59 & Nutrition-Median, Potassium-rich & 13.06 & 24 & 147\\
         \hline
         Grape & 98 & Balanced & 150 & 25 & 229\\
         \hline
         Orange Skin & 629 & Nutrition-median, Potassium-rich & 93.75 & 21 & 212\\
         \hline
         Tea Bag & 194 & Nutrition-rich, Nitrogen-rich & 4160 & 650 & 2000\\
         \hline
         Avocado Skin & 196 & Nutrition-rich, Nitrogen-rich & 1100 & 141 & 459\\
         \hline
         Chicken Bone & 161 & Nutrition-rich, Phosphorus-rich & 646.88 & 2040 & 40\\
         \hline
         Cooked Fish & 58 & Nutrition-rich, Nitrogen-rich & 2610 & 205 & 372\\
         \hline
    \end{tabular}
    \label{tab:class_vs_images}
\end{table*}

\begin{itemize}
    \item We have compiled and annotated a dataset from high-resolution images collected from the household kitchen scraps and food waste generated after meals. We further narrowed down the initially diverse food waste classes to 19 nutrition-rich categories, facilitating the estimation of final NPK values produced by the in-house composter. 

    \item We conducted a performance evaluation of four state-of-the-art semantic segmentation models on the generated food waste dataset. This benchmarking process allowed us to capture nutritional information from food waste and assess the models' effectiveness in this context.

    \item The benchmark results underscore the effectiveness of semantic segmentation methods in discerning food waste within real household kitchens. This capability facilitates the straightforward capture of NPK values associated with food waste, streamlining the process of recycling it into compost through a home composter.
\end{itemize}

\section{Related Work}

Deep neural networks have exhibited remarkable success in diverse computer vision domains, adeptly handling tasks such as image classification \cite{chen22}, object detection \cite{li21}, segmentation \cite{he21}, and recognition \cite{Yang22}. Prominent network models within this domain encompass Convolutional Neural Networks (CNNs) and Vision Transformers \cite{Chen23}\cite{Ma22}, both extensively utilized in practical applications across various domains such as multimedia \cite{xu21}, 3D vision, agriculture \cite{zhang23}, medical image analysis cite{wilson22}\cite{xiao23}, and beyond.

In recent years, these techniques have found application in discerning food images for health and nutrition-related analysis tasks. Significantly, distinctions arise based on the associated task of the dataset and the level of annotation, encompassing categorizations such as dish, ingredient, or recipe. Two noteworthy public image segmentation datasets, namely UECFoodPix and UECFoodPixComplete, offer annotations at the dish level and collectively comprise 10,000 images \cite{b3}. These datasets prove to be well-suited for fundamental dish classification tasks \cite{b4}, contributing to the progression of food-related research within the computer vision landscape.

Addressing the need for a comprehensive food image dataset, FoodSeg103 was introduced, featuring more than 104 ingredient food classes and a total of 7,118 images \cite{b5}. This dataset is distinctive in its annotation for semantic segmentation, providing detailed annotations at both the ingredient and dish levels. Notable alternatives include ETHFood101, Recipe1M, and Geo-Dish, primarily designed for dish classification and recipe generation.

Although these datasets serve as valuable benchmarks, it is essential to acknowledge that their emphasis is on entire meals depicted in images rather than on the aspect of food waste. Furthermore, images depicting food waste present a visual contrast, posing a challenge in repurposing existing datasets for this particular task. It is noteworthy that, to date, no publicly available dataset dedicated explicitly to food waste exists in the literature. This study is the first endeavor that is dedicated specifically to food waste segmentation.


\section{Dataset}


We gathered kitchen food waste from our clients and captured high-resolution images using smartphone cameras, resulting in a total of 3,128 images at a resolution of $4032 \times 3024$. Subsequently, we trained a team of engineers to manually annotate the masks for each food waste instance and assign a class name using the Labelbox platform. This effort led to the annotation of 93 distinct classes, generating 29,433 instance annotations from the captured images.

Aligned with the study's objective of identifying nutritionally rich food waste contributing to high-quality compost with NPK values, the initial set of 93 classes underwent refinement. Following guidelines from \cite{b15}, the classes were narrowed down to 15 nutrition-rich categories and 4 nutrition-light but high-performing classes. After excluding images lacking annotations for the 19 selected classes, the instance annotations were consolidated for each image to create semantic segmentation masks. This process resulted in a dataset comprising 2,912 images. Table \ref{tab:class_vs_images} listed the final 19 distinct food waste classes, together with the associated number of images and nutritional information for each class. The majority of these food waste classes exhibit nutritional richness or a balanced composition in terms of their NPK values, as stipulated by \cite{b14}, with some classes being particularly rich in specific nitrogen (N), phosphorus (P), or potassium (K) values. Please note that we also include 4 nutrition-light classes due to their prevalence in our kitchen despite their lower nutritional content.

In comparison to other publicly available food datasets, ours is the first to primarily focus on the challenge of semantic segmentation within food waste classes post-meals. In our study, we adopt 10-fold cross-validation during the experiments. For this purpose, we shuffled the dataset randomly and split all images into 10 roughly equal groups, maintaining the class distribution if possible. 

\begin{figure*}
    \centering
    \includegraphics[width=\textwidth]{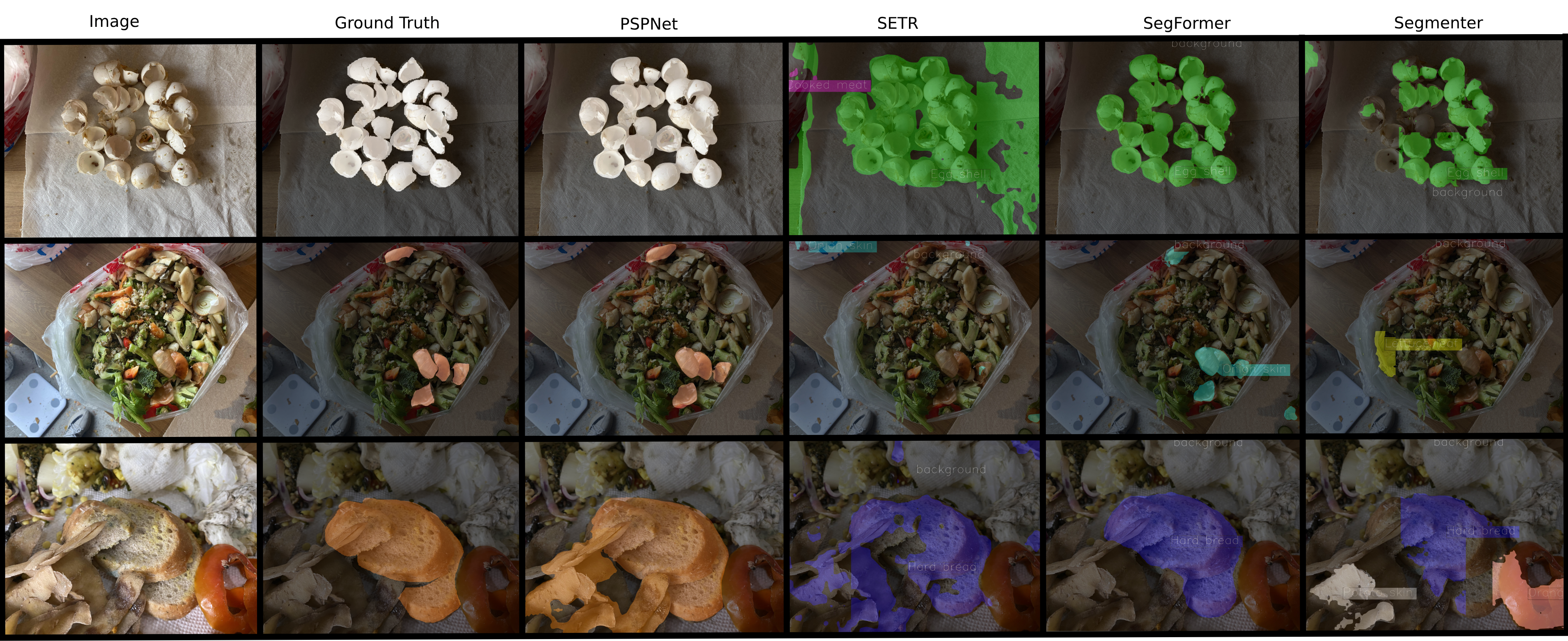}
    \caption{Visual results comparing the ground truths to their corresponding segmentation predictions are presented for cases involving egg shells, onion skins, and hard bread across various segmentation models.}
    \label{fig:qualitative-results}
\end{figure*}


\section{Models}

In recent years, deep neural networks have garnered significant success in semantic segmentation, greatly enhancing our ability to comprehensively understand images. Architectures like U-Net, PSPNet, and SegFormer have excelled in various semantic segmentation tasks across diverse domains, including health, agriculture, and autonomous vehicles \cite{b6, b7, b8}. In this paper, we investigate the efficacy of semantic segmentation in the specific context of identifying and localizing food waste. To establish a benchmark for our experiments, we employed the following four state-of-the-art segmentation models: PSPNet \cite{b9}, SETR \cite{b10}, Segmenter \cite{b11}, and SegFormer \cite{b12}.

PSPNet, introduced in \cite{b9}, is a semantic segmentation model characterized by its incorporation of a pyramid pooling module. This module enables the capture of multiscale information, enhancing segmentation accuracy by fostering context awareness. PSPNet has gained widespread popularity as a segmentation network, demonstrating state-of-the-art results across diverse computer vision applications. Its success extends to domains such as autonomous vehicles, medical image analysis, and other tasks requiring complex scene understanding.

The Segmentation Transformer (SETR) \cite{b10}, is a trailblazer in leveraging transformer architecture for segmentation tasks. This model seamlessly integrates convolutional layers with transformer layers to extract features from image patches. The initial convolutional layers focus on capturing low-level features, while the transformer layers adeptly handle high-level semantic information and context. SETR adopts a hybrid architecture, demonstrating its efficacy through remarkable results in diverse segmentation benchmarks.

Segmenter \cite{b11} is a recent transformer model designed for semantic segmentation. Unlike SETR, this model is entirely transformer-based and adopts an encoder-decoder architecture. It maps a sequence of patch embeddings to pixel-level class annotations. The transformer encoder processes the sequence of patches, followed by decoding through either a point-wise linear mapping or a mask transformer. Notably, the Segmenter model excels in capturing long-range dependencies and contextual information within images, showcasing its effectiveness in comprehending complex scenes and accurately segmenting objects in images.

SegFormer \cite{b12} represents an extension of the transformer architecture tailored for computer vision applications. In SegFormer, a grid of image patches is treated as a sequence of tokens, which undergo processing by transformer layers. This model innovatively combines local self-attention with global self-attention. Local self-attention is applied within image patches to capture fine-grained details, while global self-attention captures long-range dependencies across patches. To address the high computational cost associated with self-attention in large images, SegFormer incorporates efficient attention mechanisms. These mechanisms enable the model to focus on crucial image regions, thereby reducing overall computational complexity. SegFormer has showcased competitive performance across various computer vision benchmarks, owing to its efficiency and accuracy, positioning it as a promising architecture for a range of segmentation tasks.

\begin{table}
    \centering
    \caption{Results from semantic segmentation models.}
    \begin{tabular}{|| c | c | c ||} 
         \hline
         \textbf{Method} & \textbf{Backbone} & \textbf{mIoU} \\ [0.1ex]
         \hline
         PSPNet & ResNet50-D8 & $57.91 \pm 2.59$ \\
         \hline
         SegFormer & MIT-B0 & $65.78 \pm 3.22$ \\
         \hline
         \textbf{SegFormer} & \textbf{MIT-B5} & $\textbf{67.08} \pm \textbf{3.25}$ \\
         \hline
         SETR\_naive & ViT-L & $40.45 \pm 3.34$ \\
         \hline
         Segmenter\_mask & ViT-B\_16 & $45.16 \pm 3.71$ \\
         \hline
         \end{tabular}
    \label{tab:segresults}
\end{table}

\begin{table*}[t]
    \centering
    \caption{Class-based results from the semantic segmentation models.}
    \begin{tabular}{|| c | c | c | c | c ||} 
         \hline
         \multirow{2}{*}{} & \multicolumn{4}{| c ||}{Models} \\
         \textbf{Class} & \textbf{PSPNet} & \textbf{SegFormer} & \textbf{SETR} & \textbf{Segmenter} \\ [0.5ex]
         \hline
         Banana Skin & $70.51 \pm 1.95$ & $72.07 \pm 0.41$ & $48.31 \pm 5.87$ & $63.19 \pm 3.68$ \\
         \hline
         Egg Shell & $73.65 \pm 2.05$ & $74.99 \pm 3.72$ & $31.12 \pm 2.45$ & $58.65 \pm 7.21$ \\
         \hline
         Lettuce Leaf & $49.19 \pm 9.58$ & $57.89 \pm 1.42$ & $38.56 \pm 4.93$ & $39.33 \pm 4.27$ \\
         \hline
         Hard Bread & $69.51 \pm 6.36$ & $81.86 \pm 3.98$ & $55.56 \pm 3.23$ & $73.43 \pm 2.38$ \\
         \hline
         Cooked Meat & $56.61 \pm 4.72$ & $44.25 \pm 7.93$ & $38.46 \pm 8.02$ & $26.71 \pm 5.36$ \\
         \hline
         Onion Skin & $52.14 \pm 5.55$ & $57.32 \pm 4.70$ & $32.22 \pm 3.17$ & $39.16 \pm 4.87$ \\
         \hline
         Potato Skin & $33.63 \pm 7.25$ & $30.37 \pm 8.34$ & $30.89 \pm 5.96$ & $21.38 \pm 10.02$ \\
         \hline
         Apple Core & $58.93 \pm 8.29$ & $74.10 \pm 4.70$ & $32.83 \pm 5.34$ & $35.61 \pm 11.14$ \\
         \hline
         Orange & $51.01 \pm 20.28$ & $68.51 \pm 4.48$ & $63.34 \pm 6.07$ & $68.17 \pm 1.35$ \\
         \hline
         Waffle & $62.24 \pm 24.38$ & $88.31 \pm 4.51$ & $72.46 \pm 20.35$ & $22.34 \pm 23.72$ \\
         \hline
         Apple Peel & $40.82 \pm 6.73$ & $56.05 \pm 10.91$ & $20.07 \pm 19.97$ & $37.21 \pm 8.11$ \\
         \hline
         Corn Leaves & $68.17 \pm 12.72$ & $86.89 \pm 2.61$ & $44.58 \pm 10.07$ & $63.36 \pm 8.05$ \\
         \hline
         Cucumber & $65.64 \pm 24.07$ & $75.67 \pm 5.31$ & $63.58 \pm 12.75$ & $74.79 \pm 2.47$ \\
         \hline
         Grape & $64.41 \pm 7.89$ & $75.61 \pm 7.67$ & $65.72 \pm 6.71$ & $56.65 \pm 5.24$ \\
         \hline
         Orange Skin & $51.39 \pm 4.38$ & $56.23 \pm 3.93$ & $40.88 \pm 3.84$ & $50.96 \pm 2.93$ \\
         \hline
         Tea Bag & $56.19 \pm 6.43$ & $58.87 \pm 6.62$ & $23.07 \pm 6.52$ & $47.50 \pm 7.84$ \\
         \hline
         Avocado Skin & $49.16 \pm 5.75$ & $55.52 \pm 2.85$ & $28.27 \pm 14.61$ & $26.91 \pm 9.13$ \\
         \hline
         Chicken Bone & $56.13 \pm 5.61$ & $61.01 \pm 3.21$ & $42.51 \pm 4.26$ & $55.72 \pm 5.94$ \\
         \hline
         Cooked Fish & $40.83 \pm 18.41$ & $57.64 \pm 2.03$ & $49.21 \pm 8.67$ & $48.51 \pm 1.67$ \\
         \hline
    \end{tabular}
    \label{tab:classresults}
\end{table*}


\section{Experiments}

\textit{\textbf{Pre-Processing:}} The initial images collected were of dimensions $4032 \times 3024$. The dataset was partitioned randomly into 10 approximately equal groups to facilitate 10-fold cross-validation. Subsequently, all images were resized to $1024 \times 1024$ and subjected to random cropping, retaining 75\% of the image. Additionally, horizontal flipping was applied randomly with a 50\% probability, and photometric distortion was introduced before the training process. To address the substantial class imbalance within the dataset, class weights were computed as outlined below.
\begin{equation}
    W_{class\;i} = \dfrac{\textit{Total\;\;Pixels}}{\textit{Class\;i\;\;Pixels}}
\end{equation}

\textit{\textbf{Training Setup:}} The models were all implemented in PyTorch in Python using the MMSegmentation framework \cite{b13}. They were trained with 4 NVIDIA Tesla V100 GPUs with 48G memory in total. We used SETR\_Naive with a ViT-L backbone, Segmenter\_Mask with a ViT-B\_16 backbone, and SegFormer with an MIT-B0 backbone.

\textit{\textbf{Training Pipeline:}} For the PSPNet, Stochastic Gradient Descent (SGD) was used with a learning rate of 0.01, momentum of 0.9, and a weight decay of 0.0005. The model was trained with a batch size of 2 for 80,000 iterations. We adopt a polynomial learning rate decay schedule and employ SGD as the optimizer for the SETR and Segmenter models. We set the initial learning rate at 0.001. Momentum and weight decay are set to 0.9 and 0 respectively for all the experiments on both the models. For the SegFormer network, we trained the models using AdamW optimizer and a batch size of 2 for 160K iterations. The learning rate and weight decay were set to an initial value of 0.00006 and 0.01 respectively and then used a “poly” LR schedule with factor 1.0 by default.
To calculate the loss during training, we utilized Cross-Entropy Loss. All other hyperparameters were kept consistent with their default implementation.

\textit{\textbf{Model Evaluation:}} To evaluate our models, we used the intersection over union (IoU) metric, a foundational evaluation metric used in detection and segmentation.
\begin{equation}
    IoU = \frac{\textit{Area\;of\;Overlap}}{\textit{Area\;of\;Union}}
\end{equation}

\section{Results and Analysis}

In our experiments, we trained four state-of-the-art semantic segmentation models on our created food waste dataset and performed 10-fold cross-validation. The performance of the models is shown in Table \ref{tab:segresults}. The results show the mean and standard deviation of the mIoU for each model. SegFormer with an MIT-B5 backbone achieved the best result with an mIoU of $67.08 \pm 3.25$. To further delineate the segmentation results based on each specific food waste class, we also provide the class-based segmentation results in Table \ref{tab:classresults}. Out of the 19 food waste classes, egg shells and banana skins were consistently the top performing classes while potato skin, cooked fish, and apple peel were the worst performing. Waffle, cucumber, and orange classes also had the highest standard deviation, likely due to their low training samples.

A visualization of the qualitative results is presented in Fig. \ref{fig:qualitative-results}. The figure illustrates the disparities between semantic segmentation predictions and ground truths for instances involving eggshells, onion skin, and hard bread. All models demonstrate commendable performance. We can see that SegFormer and PSPNet consistently generate superior prediction masks characterized by precise boundaries, effectively avoiding confusion with visually similar objects in the images. In contrast, SETR and Segmenter exhibit challenges in distinguishing between backgrounds that share visual similarities with other classes, resulting in less distinct segmentation 


In the initial dataset, classes with high nutritional values in NPK were retained, given their significant contribution to the nutritional content of the resulting compost. However, this selective inclusion led to a class imbalance in the food waste dataset, impacting the model's performance, as evident in the class-wise results presented in Table \ref{tab:classresults}. The close relationship between image count and class performance is apparent, as reflected by the high standard deviation for classes with a relatively lower number of images in the dataset. The models tend to overfit the data due to the lower training samples, resulting in reduced generalizability. Based on our observations and the class-specific results, an optimal number of images for consistent and accurate results is estimated to fall within the range of 400 to 600 images per class.

The qualitative results indicate that classes with low standard deviation exhibit more consistent, accurate, and higher-quality mask predictions. This observation is particularly evident in classes such as eggshells, onion skin, hard bread, and banana peels. Conversely, classes like orange and orange peels, as well as apple cores and apple peels, display a substantial overlap in visual characteristics, presenting a greater challenge in distinguishing one class from the other. As a result, these classes exhibit higher difficulty levels, leading to increased variability and less precise segmentation outcomes.

\section{Conclusion}
Semantic segmentation is essential for predicting NPK values from images. In this study, we utilized high-resolution images of kitchen scraps and food waste, narrowing down the initially diverse waste classes to 19 nutrition-rich classes for experimentation. Based on the dataset, we have benchmarked four state-of-the-art semantic segmentation models by incorporating various data augmentation techniques and leveraging pre-trained weights.

Our results highlight that transformer-based models, particularly SegFormer, exhibit superior accuracy. Additionally, from qualitative assessments, PSPNet emerges as a strong choice, generating high-quality masks for classes it can effectively learn. PSPNet also demonstrates proficiency in distinguishing the background class from foreground food waste. The promising outcomes from our experiments provide a clear path for future work, wherein we aim to extrapolate NPK values for each food waste class based on their detected masks. By expanding our dataset to include more images for adequate representation of each class, we plan to explore further enhancements in semantic segmentation to achieve higher-quality masks for more accurate calculation of NPK values across various food waste classes.


\end{document}